\documentclass[sigconf]{acmart}
\usepackage{stfloats}
\usepackage{algorithm}
\usepackage{algorithmic}

\copyrightyear{2023}
\acmYear{2023}
\setcopyright{acmlicensed}\acmConference[MMAsia '23]{ACM Multimedia Asia
2023}{December 6--8, 2023}{Tainan, Taiwan}
\acmBooktitle{ACM Multimedia Asia 2023 (MMAsia '23), December 6--8, 2023,
Tainan, Taiwan}
\acmPrice{15.00}
\acmDOI{10.1145/3595916.3626448}
\acmISBN{979-8-4007-0205-1/23/12}

\begin{document}

\title{RanLayNet: A Dataset for Document Layout Detection used for Domain Adaptation and Generalization}

\author{Avinash Anand}
\email{	avinasha@iiitd.ac.in}
\affiliation{%
  \institution{IIIT, Delhi, MIDAS Lab}
  \city{New Delhi}
  \country{India}}

\author{Raj Jaiswal}
\email{Jaiswalofficial2908@gmail.com}
\affiliation{%
  \institution{IIIT, Delhi, MIDAS Lab}
  \city{New Delhi}
  \country{India}}
  
\author{Mohit Gupta}
\email{mohit22112@iiitd.ac.in}
\affiliation{%
  \institution{IIIT, Delhi, MIDAS Lab}
  \city{New Delhi}
  \country{India}}
  
\author{Siddhesh S Bangar}
\email{siddheshb008@gmail.com}
\affiliation{%
\institution{Vidyalankar Institute of Technology, Bombay, MIDAS Lab}
\city{New Delhi}
\country{India}}
  
\author{Pijush Bhuyan}
\email{pijush22049@iiitd.ac.in}
\affiliation{%
  \institution{IIIT, Delhi, MIDAS Lab}
  \city{New Delhi}
  \country{India}}

\author{Naman Lal}
\email{namanlal.lal92@gmail.com}
\affiliation{%
  \institution{IIIT, Delhi, MIDAS Lab}
  \city{New Delhi}
  \country{India}}

\author{Rajeev Singh}
\email{rs.rajeevsingh0603@gmail.com}
\affiliation{%
\institution{Vidyalankar Institute of Technology, Bombay, MIDAS Lab}
\city{New Delhi}
\country{India}}

\author{Ritika Jha}
\email{08ritikajha@gmail.com}
\affiliation{%
\institution{Jawaharlal Nehru University, Delhi, MIDAS Lab}
\city{New Delhi}
\country{India}}

\author{Rajiv Ratn Shah}
\email{rajivratn@iiitd.ac.in}
\affiliation{%
  \institution{IIIT, Delhi, MIDAS Lab}
  \city{New Delhi}
  \country{India}}

\author{Shin'ichi Satoh}
\email{satoh@nii.ac.jp}
\affiliation{%
  \institution{National Institute of Informatics}
  \city{Tokyo}
  \country{Japan}}

\renewcommand{\shortauthors}{Avinash A et al.}

\begin{abstract}
Large ground-truth datasets and recent advances in deep learning techniques have been useful for layout detection. However, because of the restricted layout diversity of these datasets, training on them requires a sizable number of annotated instances, which is both expensive and time-consuming. As a result, differences between the source and target domains may significantly impact how well these models function. To solve this problem, domain adaptation approaches have been developed that use a small quantity of labeled data to adjust the model to the target domain. In this research, we introduced a synthetic document dataset called RanLayNet, enriched with automatically assigned labels denoting spatial positions, ranges, and types of layout elements. The primary aim of this endeavor is to develop a versatile dataset capable of training models with robustness and adaptability to diverse document formats. Through empirical experimentation, we demonstrate that a deep layout identification model trained on our dataset exhibits enhanced performance compared to a model trained solely on actual documents. Moreover, we conduct a comparative analysis by fine-tuning inference models using both PubLayNet and IIIT-AR-13K datasets on the Doclaynet dataset. Our findings emphasize that models enriched with our dataset are optimal for tasks such as achieving \textbf{0.398} and \textbf{0.588} mAP95 score in the scientific document domain for the \textbf{TABLE} class. GitHub Repository: \href{https://github.com/midas-research/randomlaynet/tree/main}{\textbf{https://github.com/midas-research/randomlaynet/tree/main}}.
\end{abstract}



\maketitle
\section{Introduction}

\begin{figure*}[ht]
\centering
\Description{RanLayNet Dataset Sample}
  \includegraphics[width=0.9\linewidth]{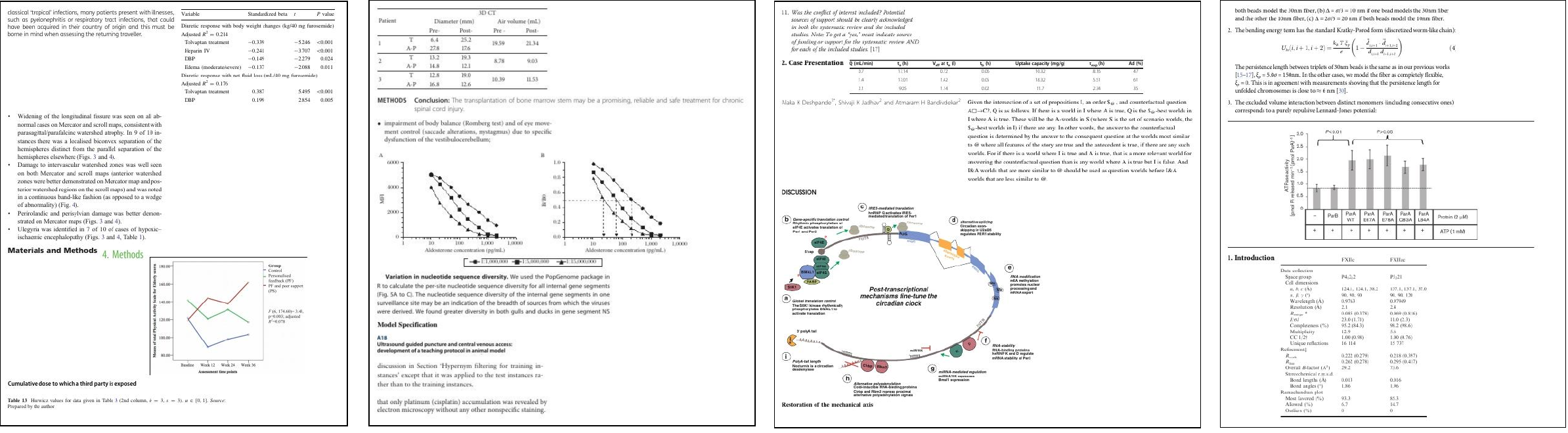}
\caption{RanLayNet Dataset Sample}
\label{random_laynet_samples}
\end{figure*}

Document layout analysis encompasses diverse structures, shapes, and appearances, further compounded by scarce annotated data and shifting domains. Growing internet accessibility floods the public domain with documents, magnifying data annotation's laboriousness. Despite a fundamental structure, documents harbor varied textual and non-textual elements with distinct attributes. The main problem is understanding how documents are laid out, which is important for getting information from them. Effective business intelligence relies on large-scale data extraction from documents to inform decisions, with layout driving extraction performance. 

The deep learning revolution, as in various fields, has similarly illuminated layout understanding challenges. Hence, annotated data is crucial for guiding learning tasks, but obtaining extensive data is challenging in real situations. Privacy concerns and manual annotation efforts hinder data availability, especially for sensitive images. Data augmentation, particularly synthetic image generation, stands out as a successful strategy to address this limitation.

Domain generalization and adaptation are distinct concepts in machine learning, primarily in supervised learning. Both aim to improve model performance on new data, but they vary in goals. Generalization involves excelling on unseen data from the same distribution as training data. Meanwhile, domain adaptation transfers knowledge from one domain to another with dissimilar distributions, aiming to enhance the model's target domain performance using labeled data from the source domain.
\begin{figure*}[bp]
\centering
\Description{Doclaynet Results on Fine-tuned RanLayNet model}
  \includegraphics[width=0.9\linewidth]{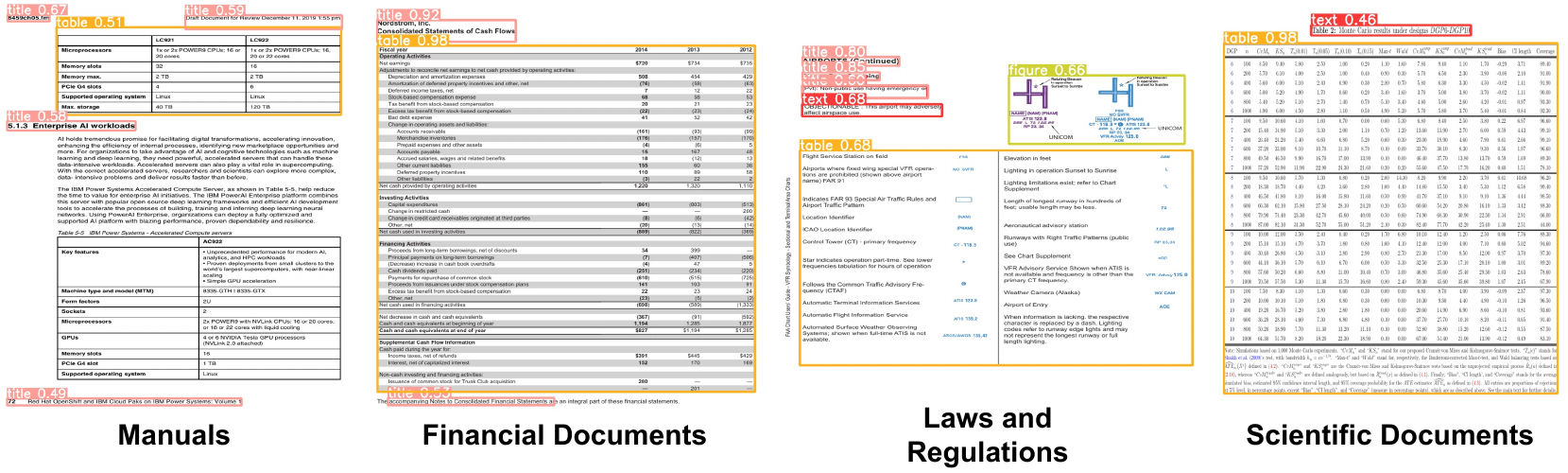}
\caption{Doclaynet Results on Fine-tuned RanLayNet model}
\label{doclaynet_inference_results}
\end{figure*}

Early research in domain adaptation for visual applications was extensively explored by Peng and Saenko et al.~\cite{peng2015domain}, who conducted a survey on various techniques. Although previous studies introduced domain adaptation methods like feature selection~\cite{zhang2022docunet}, transfer learning~\cite{li2022unsupervised}, and domain-specific knowledge integration these approaches often face constraints in scalability, performance, or domain-specific applicability. The need for an effective solution to these limitations remains significant. The main goal is to enhance the model's ability to work well across different document layouts and object detection scenarios. Although the cross-domain detection approach is useful for document analysis and object detection, the absence of appropriate transition samples between different domains poses a challenge for effectively refining the model's training to perform well on specific target samples.

In this research, we aimed to tackle these challenges by creating a new set of data using well-known benchmark datasets like Publaynet~\cite{zhong2019publaynet}. We also conducted tests on another dataset called Doclaynet~\cite{min2020doclaynet} using a carefully adjusted YOLO model~\cite{redmon2016you}. This helped us examine how well the trained model performs under different conditions and challenges.

Our contributions in this research are as follows:
\begin{enumerate}
    \item We introduced \textbf{RanLayNet} dataset, generated synthetically with automatic labeling for layout elements.
    \item The empirical tests demonstrate that a deep model trained on synthetic documents achieves comparable or marginally superior performance compared to models trained on actual documents.
    \item The models trained on the RanLayNet dataset excel in domain adaptation/generalization for layout identification.
\end{enumerate}

\section{Related Work}
In recent years, the field of domain adaptation~\cite{khodabandeh2019robust} has witnessed notable advancements, addressing the challenges of knowledge transfer from a source domain with a different data distribution to target domain. One note-worthy contribution is the work of Zhang et al.~\cite{zhong2021does}, who introduced a progressive self-training framework for unsupervised domain adaptation. Their method gradually makes use of both labeled source domain data and unlabeled target domain data, iteratively refining the model's predictions and incorporating target domain samples. 

Early methodologies in document layout analysis relied on rule-based algorithms and heuristic approaches \cite{ahmad2016information}. However, the contemporary landscape has witnessed a shift towards addressing this issue through deep learning techniques. They involve the utilization of object detection models \cite{girshick2014rich,girshick2015fast,ren2015faster,dai2021dynamic,jocher2021ultralytics,carion2020end,tan2020efficientdet,anand2024tcocr}, which have demonstrated substantial enhancements in accuracy and speed over the past decade. Notably, most cutting-edge object detection methods can be easily trained and implemented with minimal effort due to standardized ground-truth data formats~\cite{lin2014microsoft} and the widespread adoption of common deep learning frameworks \cite{wu2019detectron2}. Reference datasets such as PubLayNet \cite{zhong2019publaynet} and DocBank \cite{li2020docbank} conveniently present their data in the widely accepted COCO format \cite{lin2014microsoft}. 
\section{Datasets}

\begin{figure*}[bp]
\centering
\Description{Pipeline of RanLayNet generation. Crops of PubLayNet are randomly pasted on white canvas on the basis of remaining spaces on the canvas.}
  \includegraphics[width=0.8\linewidth]{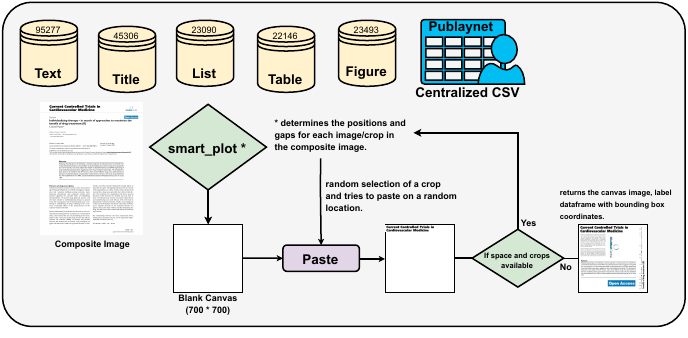}
\caption{Pipeline of RanLayNet generation. Crops of PubLayNet are randomly pasted on white canvas on the basis of remaining spaces on the canvas.}
\label{pipeline}
\end{figure*}

\subsection{Publaynet}
To create a high-quality document layout dataset called PubLayNet, Zhong et al.~\cite{zhong2019publaynet} suggest a technique for automatically annotating the document layout of more than 1 million PubMed Central PDF publications. The collection includes more than 360k samples of pages and includes common document layout components like text, titles, lists, figures, and tables. They demonstrate that a model can be trained on an automatically annotated dataset to recognize the format of scientific papers and that the model that has already been trained on the dataset can serve as a more solid foundation for transfer learning.

\subsection{IIIT-AR-13K}
Mondal et al.~\cite{mondal2020iiitar13k} introduced IIIT-AR-13K, a dataset designed for the localization of graphical objects within the annual reports of diverse companies, which are a distinct category of business documents.They randomly choose publicly accessible annual reports in English and various other languages, including French, Japanese, Russian, among others, from over ten years of twenty-nine distinct companies. They perform manual annotation of bounding boxes for five distinct categories of graphical elements that are commonly present in annual reports: tables, figures, natural images, logos, and signatures. The IIIT-AR-13K dataset comprises 13,000 annotated pages, encompassing 16,000 tables, 3,000 figures, 3,000 natural images, 500 logos, and 600 signatures.

\subsection{Doclaynet}
Min et al.~\cite{min2020doclaynet} introduce the DocLayNet dataset, a comprehensive resource for page-by-page layout annotation. Encompassing 80863 distinct document pages, the dataset offers precise bounding-box annotations for 11 distinct class labels, elucidating layout attributes. Notably, a portion of these pages is enriched with double or triple annotations, totaling 91104 instances. The annotations furnish intricate layout insights, employing designated labels such as Caption, Footnote, Formula, List-item, Page-footer, Page-header, Picture, Section-header, Table, Text, and Title.

\subsection{RanLayNet}
The RanLayNet dataset introduces higher variability in document layouts, surpassing existing datasets by presenting complex structures. It covers diverse layout classes, offering a balanced representation of elements. This variety equips models to handle unpredictable real-world layouts. Deep learning models for document layout learn from RanLayNet, gaining insights into spatial positions, extents, and categories of elements. Exposure to diverse layouts helps these models adapt to real-world variations. The dataset's generality enables training models for various layouts, addressing domain shifts.

\begin{table}[ht]
    \centering
    \caption{Class Label Count \& Distribution of the RanLayNet Dataset.}
    \label{tab:stats_randomlaynet}
    \setlength{\tabcolsep}{3.8\tabcolsep}
    \begin{tabular}{|l|c|c|}
         \hline
         \textbf{Label} & \textbf{Count} & \textbf{Percentage (\%)} \\
         \hline
         \hline
         \textbf{Text} & 95,227 & 45.52 \\
         \hline
         \textbf{Title} & 45,306 & 21.65 \\
         \hline
         \textbf{List} & 23,090 & 11.03 \\
         \hline
         \textbf{Table} & 22,146 & 10.58 \\
         \hline
         \textbf{Figure} & 23,493 & 11.22 \\
         \hline
     \end{tabular}
\end{table}

\begin{table*}[ht]
\centering
\caption{Inference of "TABLE" Class on Doclaynet Documents}
\label{tab:my_label}
{\renewcommand{\arraystretch}{1.2}
\setlength{\tabcolsep}{2\tabcolsep}
\begin{tabular}{|l|c|c|c|c|c|c|c|c|}
\hline
\textbf{Domain} & \multicolumn{4}{|c|}{\textbf{IIIT-AR-13K}} & \multicolumn{4}{|c|}{\textbf{IIIT-AR-13K + RanLayNet}} \\
\hline
& Precision & Recall & mAP50 & mAP95 & Precision & Recall & mAP50 & mAP95 \\
\hline
Manuals & 0.194 & 0.0353 & 0.0587 & 0.0262 & \textbf{0.446} & \textbf{0.356} & \textbf{0.336} & \textbf{0.226} \\
\hline
Financial Documents & 0.00405 & 0.013 & 0.00028 & 0.00088 & \textbf{0.374} & \textbf{0.334} & \textbf{0.298} & \textbf{0.187} \\
\hline
Laws and Regulations & 0.194 & 0.353 & 0.0587 & 0.0262 & \textbf{0.619} & \textbf{0.263} & \textbf{0.293} & \textbf{0.222} \\
\hline
Scientific Documents & 0.0112 & 0.0042 & 0.00582 & 0.00219  & \textbf{0.187} & \textbf{0.474} & \textbf{0.334} & \textbf{0.398} \\
\hline
\end{tabular}}
\end{table*}

\begin{table*}[ht]
\centering
\caption{Inference of "TABLE" Class on Doclaynet Documents}
\label{tab:my_label2}
{\renewcommand{\arraystretch}{1.2}
\setlength{\tabcolsep}{2\tabcolsep}
\begin{tabular}{|l|c|c|c|c|c|c|c|c|}
\hline
\textbf{Domain} & \multicolumn{4}{|c|}{\textbf{PubLayNet}} & \multicolumn{4}{|c|}{\textbf{PubLayNet + RanLayNet}} \\
\hline
& Precision & Recall & mAP50 & mAP95 & Precision & Recall & mAP50 & mAP95 \\
\hline
Manuals & 0.488 & 0.582 & 0.421 & 0.220 & \textbf{0.562} & \textbf{0.807} & \textbf{0.761} & \textbf{0.588} \\
\hline
Financial Documents & 0.350 & 0.452 & 0.288 & 0.163 & \textbf{0.427} & \textbf{0.555} & \textbf{0.465} & \textbf{0.293} \\
\hline
Laws and Regulations & 0.039 & 0.356 & 0.337 & 0.194 & \textbf{0.294} & \textbf{0.572} & \textbf{0.350} & \textbf{0.282} \\
\hline
Scientific Documents & 0.366 & 0.621 & 0.562 & 0.376  & \textbf{0.562} & \textbf{0.807} & \textbf{0.761} & \textbf{0.588} \\
\hline
\end{tabular}}
\end{table*}

RanLayNet's diverse structures combat overfitting, fostering adaptable representations that counter domain disparities. Models trained on RanLayNet surpass those on PublayNet, showcasing robustness and adaptability to various layouts, reinforcing domain adaptation. The class label count and their distribution of RanLayNet is shown in Table.~\ref{tab:stats_randomlaynet}.

\section{Methodology}
\textbf{Problem Statement: } The objective of this project is to create a dataset that can facilitate the development of highly resilient and versatile models. These models, trained on the dataset, should possess the ability to effectively handle various structures and formats. Consequently, a model trained on this dataset would be capable of efficiently processing data in a wide range of formats.

We tackle the challenge of limited model generalization across diverse domains in domain adaptation through the utilization of 'RanLayNet'. This innovative approach involves a noisy dataset, which enhances versatility while minimizing bias.

Our approach begins by utilizing the source dataset Publaynet, which consists of fixed labels - Text, Title, List, Figure, and Table. This dataset is employed for the initial training of our model. Subsequently, a target dataset is meticulously curated, characterized by an expanded class structure and diverse domains. The central challenge in domain adaptation lies in addressing the disparities between the source and target datasets.

\begin{table}[ht]
    \centering
    \caption{IIIT-AR-13K fine-tuned on YOLO v8}
    \label{tab:table2}
    \setlength{\tabcolsep}{\tabcolsep}
    \begin{tabular}{|l|c|c|c|c|}
         \hline
         \textbf{Class} & \textbf{Precision} & \textbf{Recall} & \textbf{mAP50} & \textbf{mAP95} \\
         \hline
         \hline
         \textbf{All} & 0.946 & 0.922 & 0.957 & 0.81 \\
         \hline
         \textbf{Table} & 0.984 & 0.988 & 0.992 & 0.945 \\
         \hline
         \textbf{Figure} & 0.951 & 0.869 & 0.945 & 0.763 \\
         \hline
         \textbf{Natural Image} & 0.95 & 0.936 & 0.968 & 0.919 \\
         \hline
         \textbf{Signature} & 0.981 & 0.97 & 0.992 & 0.763 \\
         \hline
         \textbf{Logo} & 0.863 & 0.845 & 0.89 & 0.661 \\
         \hline
     \end{tabular}
\end{table}

\begin{table}[ht]
    \centering
    \caption{PubLayNet fine-tuned on YOLO v8}
    \label{tab:table5}
    \setlength{\tabcolsep}{1.6\tabcolsep}
    \begin{tabular}{|l|c|c|c|c|}
         \hline
         \textbf{Class} & \textbf{Precision} & \textbf{Recall} & \textbf{mAP50} & \textbf{mAP95} \\
         \hline
         \hline
         \textbf{All} & 0.910 & 0.880 & 0.914 & 0.855 \\
         \hline
         \textbf{Text} & 0.933 & 0.785 & 0.924 & 0.870 \\
         \hline
         \textbf{Title} & 0.843 & 0.958 & 0.886 & 0.705 \\
         \hline
         \textbf{List} & 0.914 & 0.780 & 0.895 & 0.832 \\
         \hline
         \textbf{Table} & 0.955 & 0.942 & 0.988 & 0.951 \\
         \hline
         \textbf{Figure} & 0.933 & 0.884 & 0.931 & 0.894 \\
         \hline
     \end{tabular}
\end{table}

RanLayNet breaks from conventional datasets by embracing diverse layout configurations, fostering adaptability, and reducing bias. It's a vital asset for domain adaptation. Starting with the source dataset and its fixed five labels, we initiate model training. This baseline performance serves to reveal limitations in adapting to the target dataset's diverse domains and extended class structure. Model evaluation comes next, as we test inference on the target dataset, revealing gaps in generalization. An essential aspect of this approach is the gradual expansion of the RanLayNet dataset. We methodically add diverse layout structures and distinct data distributions as patches onto the dataset canvas. This augmentation boosts dataset complexity and variance, exposing the model to a wider range of scenarios.

The process of creating a RanLayNet dataset with consolidated images and corresponding label information involves several steps. Initially, a .csv containing details about the images and their associated bounding boxes is provided. Leveraging the $smart\_plot$ function, positions and gaps are determined to accommodate each image or crop effectively within a composite image.

Upon successful position and gap calculation, a blank canvas is generated as the base for the composite image. Subsequently, the function iterates through each determined position. Based on the provided data, it selectively crops specific regions from the images or pastes entire images onto the composite canvas at the designated positions. As this process unfolds, the coordinates of bounding boxes are updated, and these new coordinates are stored in a label file. Iterating through all positions, the function orchestrates the arrangement and placement of images or crops onto the composite canvas, creating a comprehensive representation. Ultimately, the output comprises the composite image itself, and a label file containing adjusted bounding box information. This holistic approach streamlines the compilation of a RanLayNet dataset that offers a consolidated view of images alongside their corresponding labels, facilitating streamlined analysis and evaluation. The visual representation of the whole pipeline is shown in Fig.~\ref{pipeline}.

\section{Experiments}
Our experimental setup involves fine-tuning YOLOv8 across various datasets. We fine-tuned YOLOv8 for 40 epochs, having the batch size of 16, and num\_workers set to 4, with SGD optimizer, and a learning rate of 1e-3. For computational support, we utilized an NVIDIA RTX A6000 GPU with 512 GB of RAM. Table.~\ref{tab:table2}, ~\ref{tab:table5}, ~\ref{tab:table3}, and~\ref{tab:table4} shows that our model is able to learn the different classes during the training. In our context, \textbf{Noise Labeling}\footnote{Definition can be found in the appendix section.} is applied to a document layout identification model, aiming to enhance the model's capability in accurately detecting various elements within diverse types of documents spanning different domains. The observation from our experimentation reveals that the "Table" class demonstrates the highest mean average precision, suggesting its significance and potential impact across various applications.

To leverage this observation, we fine-tune YOLOv8 models using both a clean source dataset (IIIT-AR-13K and Publaynet) and later we further fine-tuned both model on our dataset. This fine-tuning process strategically aims to enhance the model's performance across multiple metrics and diverse domains, thereby showcasing the achieved adaptability and generalization capabilities through this approach.

\begin{table}[ht]
    \centering
    \caption{RanLayNet fine-tuned on YOLO v8 (using the weights of IIIT-AR-13k)}
    \label{tab:table3}
    \setlength{\tabcolsep}{1.6\tabcolsep}
    \begin{tabular}{|l|c|c|c|c|}
         \hline
         \textbf{Class} & \textbf{Precision} & \textbf{Recall} & \textbf{mAP50} & \textbf{mAP95} \\
         \hline
         \hline
         \textbf{All} & 0.968 & 0.945 & 0.974 & 0.918 \\
         \hline
         \textbf{Text} & 0.958 & 0.935 & 0.978 & 0.927 \\
         \hline
         \textbf{Title} & 0.949 & 0.981 & 0.98 & 0.8 \\
         \hline
         \textbf{List} & 0.956 & 0.908 & 0.971 & 0.944 \\
         \hline
         \textbf{Table} & 0.988 & 0.988 & 0.994 & 0.99 \\
         \hline
         \textbf{Figure} & 0.989 & 0.912 & 0.947 & 0.931 \\
         \hline
     \end{tabular}
\end{table}

\begin{table}[ht]
    \centering
    \caption{RanLayNet fine-tuned on YOLO v8 (using the weights of PubLayNet)}
    \label{tab:table4}
    \setlength{\tabcolsep}{1.6\tabcolsep}
    \begin{tabular}{|l|c|c|c|c|}
         \hline
         \textbf{Class} & \textbf{Precision} & \textbf{Recall} & \textbf{mAP50} & \textbf{mAP95} \\
         \hline
         \hline
         \textbf{All} & 0.925 & 0.884 & 0.934 & 0.858 \\
         \hline
         \textbf{Text} & 0.946 & 0.794 & 0.948 & 0.873 \\
         \hline
         \textbf{Title} & 0.843 & 0.966 & 0.903 & 0.711 \\
         \hline
         \textbf{List} & 0.922 & 0.795 & 0.896 & 0.842 \\
         \hline
         \textbf{Table} & 0.969 & 0.965 & 0.985 & 0.958 \\
         \hline
         \textbf{Figure} & 0.947 & 0.900 & 0.940 & 0.904 \\
         \hline
     \end{tabular}
\end{table}

By intentionally introducing noisy labels, we can study the behavior of models under different types of noisy conditions, and develop strategies for building more robust and accurate models. It can also help identify the types of errors or biases that are more common in the data, and develop methods to address them. Earlier research in robust learning \cite{khodabandeh2019robust} primarily concentrated on image classification scenarios with a limited number of separate classes. Initial studies employed instance-independent noise models, wherein each class was susceptible to confusion with other classes, irrespective of the content of the instance. More recently, the academic discourse has transitioned towards the prediction of instance-specific label noise. To the best of our knowledge, ours is the first suggestion for a robust label noise-resistant document layout identification model.

Tables~\ref{tab:table2}, ~\ref{tab:table5}~\ref{tab:table3}, and~\ref{tab:table4} depict that the class label \textbf{"Table"} exhibits the highest mean average precision compared to the other classes. Based on this observation, we conducted \textbf{"Table"} class detection using two fine-tuned models from source datasets: IIIT-AR-13K and IIIT-AR-13K with RanLayNet. We aimed to detect "Tables" in diverse domain documents, including Manuals, Financial Documents, Laws \& Regulations, and Scientific Documents. The results presented in Table~\ref{tab:my_label}, and~\ref{tab:my_label2} demonstrate that fine-tuning the source dataset with our noisy dataset significantly enhances model performance across various metrics in different target domains. This finding provides further evidence supporting the effectiveness of our approach in enhancing adaptability and generalization capabilities.

\section{Conclusion} 
In conclusion, our proposed approach has shed light on the challenges of bridging the gap between source and target datasets with distinct label structures. The model trained on a source dataset, defined by a specific set of five labels, encountered limitations when applied to a target dataset encompassing additional classes. The discrepancy between these datasets hindered the model's generalization, prompting the exploration of innovative solutions. In response to this, we introduced the concept of \textbf{"RanLayNet"} a dynamic and unbiased approach to dataset creation. This noisy dataset, designed without a predetermined layout structure, demonstrated inherent adaptability to diverse layout configurations. This adaptability is the key to mitigating bias and promoting versatility, enabling the model to better navigate the intricate domain landscape. By leveraging the power of RanLayNet and embracing a generality paradigm, our approach has the potential to revolutionize domain adaptation strategies. Through the amalgamation of innovative methodologies, we strive to equip models with enhanced adaptability, ensuring they transcend the constraints of source-target alignment paradigms. This pursuit not only holds promise for achieving groundbreaking results but also contributes significantly to advancing the field of domain adaptation in a dynamic and evolving data landscape.

\section{Future Scope}
The current implementation, featuring five labels, showcased promising adaptability on the target domain. However, our trajectory envisions a progressive evolution, with plans to introduce diverse patches on our canvas. This expansion is anticipated to facilitate seamless generalization across a broader spectrum of domains, even without explicit label generation for both the source and target datasets.

\section{Acknowledgement}
Dr. Rajiv Ratn Shah is partly supported by the Infosys Center for AI, the Center of Design and New Media, and the Center of Excellence in Healthcare at Indraprastha Institute of Information Technology, Delhi. We sincerely appreciate the guidance and unwavering support provided by Dr. Astha Verma throughout our research. We are grateful for their time, dedication, and willingness to share knowledge, which significantly contributed to the completion of this work.

\bibliographystyle{ACM_Reference_Format.bst}
\bibliography{software}

\newpage
\appendix
\section{Appendix}
\textbf{Noise labeling}, refers to the process of intentionally adding noisy labels to the training data in order to study the effect of noisy labels on the performance of models. This is often done to evaluate the robustness of machine learning algorithms to different types and levels of label noise. 

\begin{figure*}[bp]
\centering
\Description{YOLOv8 Training/Validation Loss Curves for IIIT-AR-13k Dataset}
  \includegraphics[width=\linewidth]{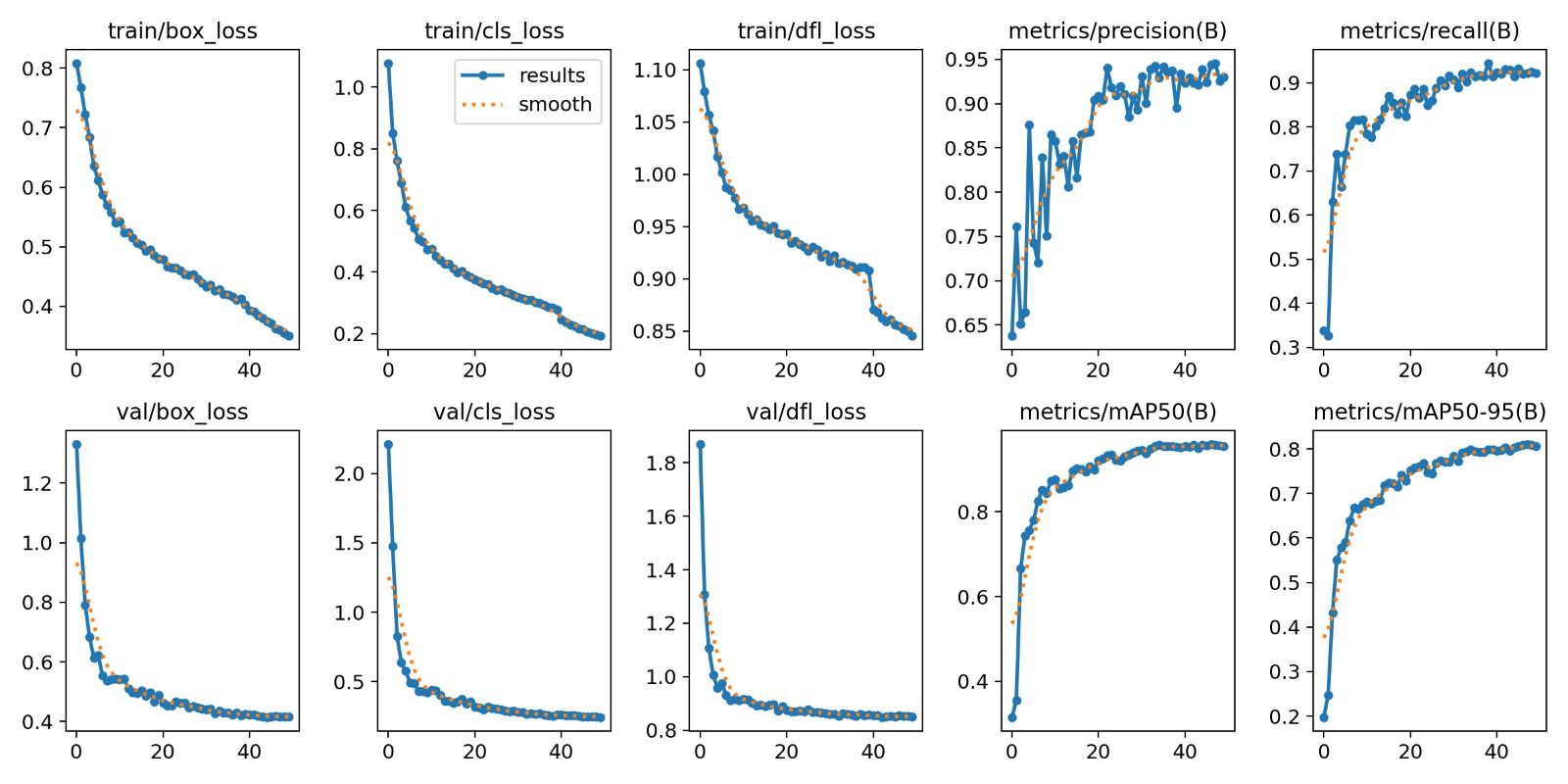}
\caption{YOLOv8 Training/Validation Loss Curves for IIIT-AR-13k Dataset}
\label{iiitd_loss_curve}
\end{figure*}

\begin{figure*}[bp]
\centering
\Description{YOLOv8 Training/Validation Loss Curves for PubLayNet}
  \includegraphics[width=\linewidth]{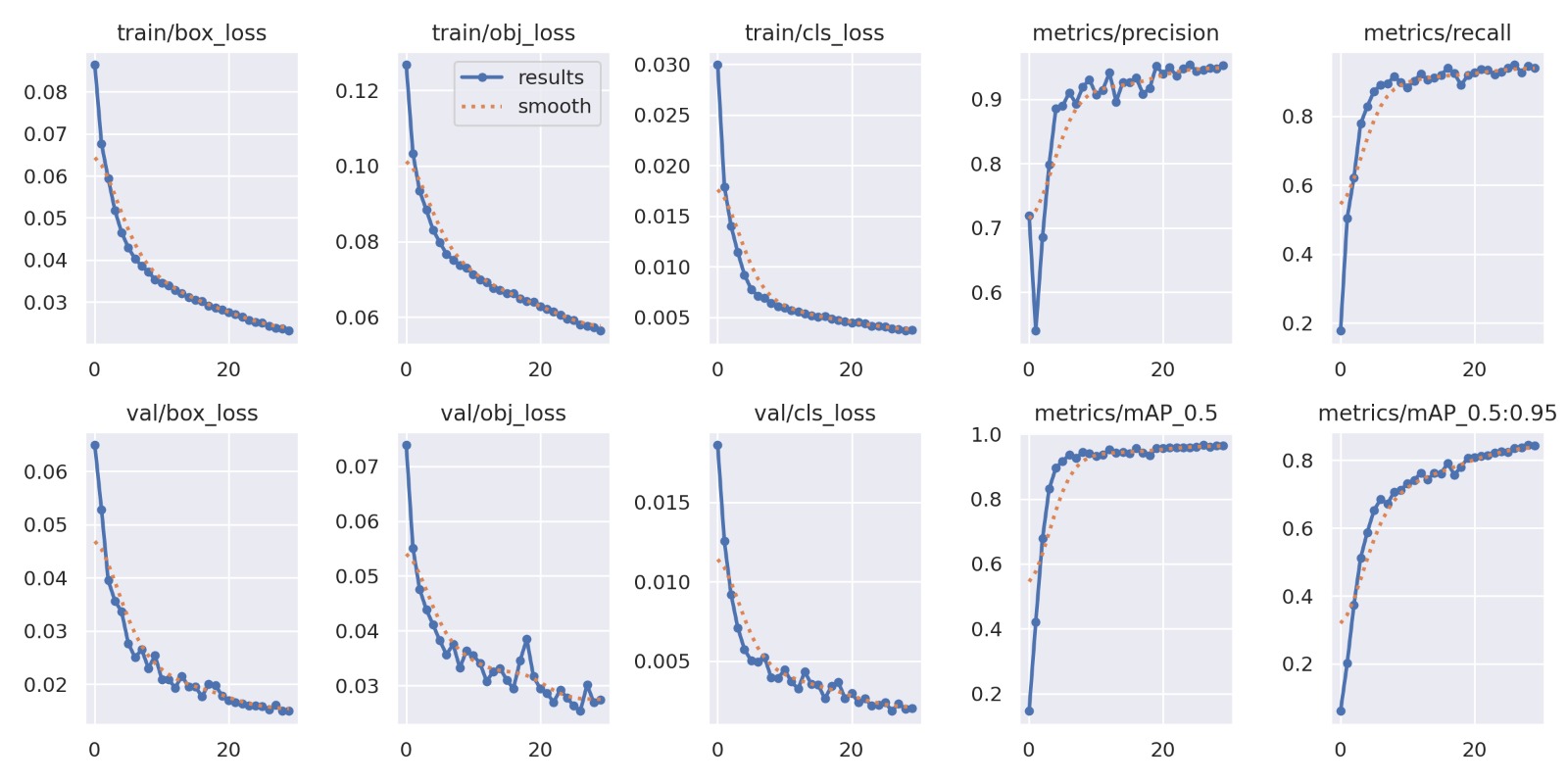}
\caption{YOLOv8 Training/Validation Loss Curves for PubLayNet}
\label{publaynet}
\end{figure*}

\begin{figure*}[bp]
\centering
\Description{YOLOv8 Training/Validation Loss Curves for RanLayNet}
  \includegraphics[width=\linewidth]{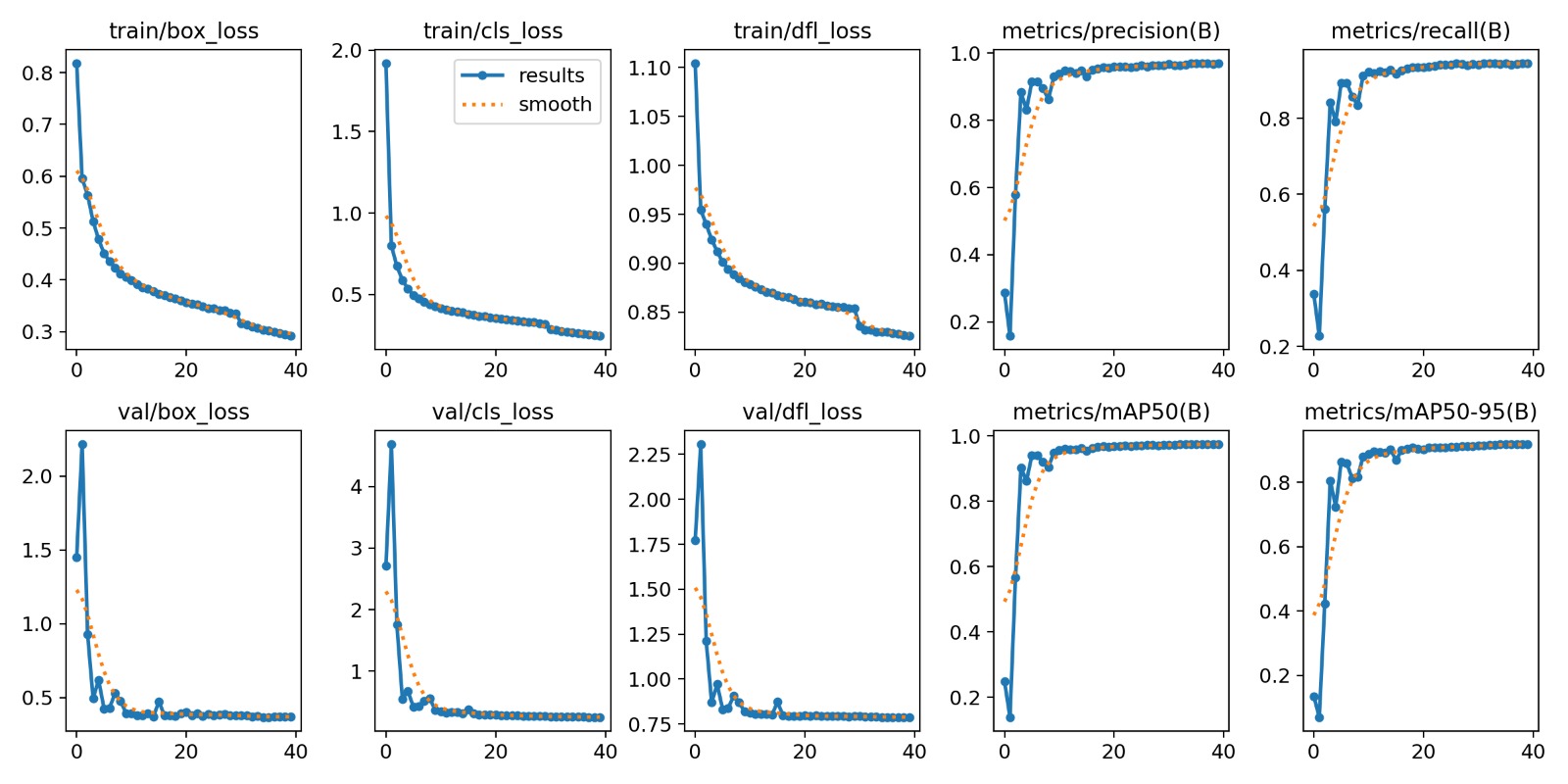}
\caption{YOLOv8 Training/Validation Loss Curves for RanLayNet}
\label{random_laynet}
\end{figure*}

The training/validation curves for YOLOv8 measures on IIIT-AR-13K, PubLayNet, and RanLayNet are in Fig.~\ref{iiitd_loss_curve}, Fig.~\ref{publaynet}, and Fig.~\ref{random_laynet} respectively. The \textbf{training loss curve} depicts improved performance over epochs, signifying effective learning and adaptation.

The \textbf{validation loss curve} evaluates generalization by reducing loss on separate data. Convergence of both curves indicates successful training, reflecting enhanced object detection accuracy. This understanding guided our approach, yielding a proficient model for real-world scenarios.

\end{document}